\documentclass[10pt,twocolumn,letterpaper]{article}

\usepackage[pagenumbers]{iccv} 

\definecolor{iccvblue}{rgb}{0.21,0.49,0.74}
\usepackage[pagebackref,breaklinks,colorlinks,allcolors=iccvblue]{hyperref}
\usepackage{arydshln}
\usepackage{adjustbox}
\usepackage{multirow}
\usepackage{float} 
\makeatletter
\newcommand{\nonumfootnote}[1]{%
  \begingroup
  \renewcommand{\thefootnote}{}%
  \def\@makefntext##1{\noindent##1}%
  \footnotetext{#1}%
  \endgroup
}
\makeatother
\def\paperID{14427} 
\def\confName{ICCV}
\def\confYear{2025}

\title{Disentangling Instruction Influence in Diffusion Transformers for Parallel
Multi-Instruction-Guided Image Editing}

\author{
    Hui Liu\textsuperscript{1,2*} \quad
    Bin Zou\textsuperscript{2,3*} \quad
    Suiyun Zhang\textsuperscript{2} \quad
    Kecheng Chen\textsuperscript{1} \quad
    Rui Liu\textsuperscript{2\S} \quad
    Haoliang Li\textsuperscript{1$\dagger$} \\[5pt]
    $^1$City University of Hong Kong \qquad
    $^2$Huawei Noah’s Ark Lab \qquad
    $^3$The University of Hong Kong  \\[8pt]
}

\begin{document}
\twocolumn[{
\maketitle
\centering
    \begin{tabular}{ccc}
    \multicolumn{3}{c}{\includegraphics[width=0.9\linewidth]{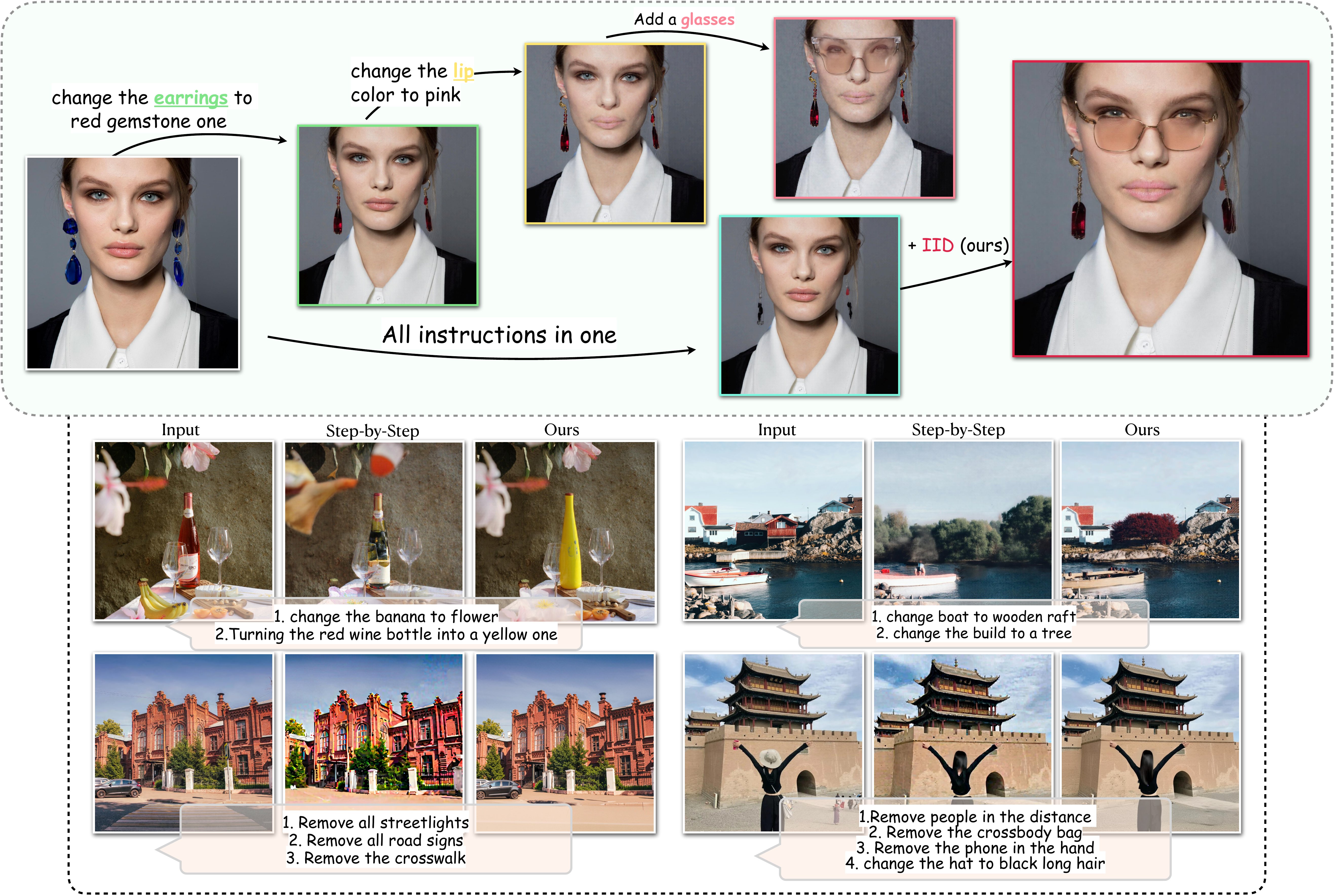}} \\
    \end{tabular}
  \captionof{figure}{
Comparison of our proposed Instruction Influence Disentanglement (IID) framework with step-by-step editing and compositing all instructions into a single one for multi-instruction image editing.}
  \label{fig1}
 \vspace*{0.4cm}
}]
\begin{abstract}
\nonumfootnote{\noindent* Equal contribution \quad \S~Project leader \quad$\dagger$  Corresponding author}
Instruction-guided image editing enables users to specify modifications using natural language, offering more flexibility and control. Among existing frameworks, Diffusion Transformers (DiTs) outperform U-Net-based diffusion models in scalability and performance. However, while real-world scenarios often require concurrent execution of multiple instructions, step-by-step editing suffers from accumulated errors and degraded quality, and integrating multiple instructions with a single prompt usually results in incomplete edits due to instruction conflicts. We propose Instruction Influence Disentanglement (IID), a novel framework enabling parallel execution of multiple instructions in a single denoising process, designed for DiT-based models. By analyzing self-attention mechanisms in DiTs, we identify distinctive attention patterns in multi-instruction settings and derive instruction-specific attention masks to disentangle each instruction’s influence. These masks guide the editing process to ensure localized modifications while preserving consistency in non-edited regions.
Extensive experiments on open-source and custom datasets demonstrate that IID reduces diffusion steps while improving fidelity and instruction completion compared to existing baselines. The codes will be publicly released upon the acceptance of the paper.


\end{abstract}


\section{Introduction}
\label{sec:intro}
Instruction-guided image editing~\citep{instructpix2pix,mutlireward,emudit,I2EBench,magicbrush} has gained increasing attention for its ability to enable users to specify editing objectives using natural language. Among various frameworks, Diffusion Transformers (DiTs)~\citep{DiT,mmDiT} exhibit superior scalability, with performance improving as model and dataset sizes increase, as seen in Omnigen~\citep{omnigen} and FluxEdit~\citep{flux_edit}, surpassing U-Net-based diffusion models~\citep{instructpix2pix, huang2025paralleledits}. However, real-world scenarios often require applying multiple modifications concurrency to an image~\cite{guo2023focus, huang2025paralleledits, I2EBench,multi1}. Given DiTs’ excellent performance in single-instruction-guided image editing tasks, extending them to multi-instruction editing is a natural yet challenging research direction.

Although exploiting an editing model step by step to each instruction~\citep{magicbrush, I2EBench, seq1} or merging multiple instructions into a single composite one may seem viable, both approaches have significant limitations. As illustrated in Fig. \ref{fig1}, the sequential execution often leads to progressive distortions (e.g., the glasses and facial features in the first-row subfigure) and degradation (e.g., the blurred background in the second subfigure of the second row)  as the number of instruction executions increases. This deterioration likely stems from cumulative errors introduced by repeated denoising processes, which disrupt the natural diffusion feature space~\cite{schedule,emudit}. Similarly, the compositional approach often fails to execute all intended modifications, typically applying only one successfully. This issue may arise from the dual influence of instructions in the editing process, ensuring that the specified modifications are applied to target regions while preserving the integrity of unedited areas. When multiple instructions are processed simultaneously, the model tends to prioritize one over the others, reducing their effectiveness for maintaining the integrity of unedited regions (termed instruction conflicts).


Intuitively, disentangling the influence of multiple instructions to ensure each one only affects its target region can mitigate conflicts, thus enabling parallel image editing with various instructions in a single pass. However, no existing methods address this challenge for DiT-based image editing models. Prior research~\citep{guo2023focus, huang2025paralleledits,xie2023boxdiff,hertz2022prompt} has focused on U-Net-based diffusion models, leveraging cross-attention maps between edited object tokens and noised images to generate masks to guide attention computations. These techniques, however, are not directly applicable to DiTs due to fundamental architectural differences, namely the replacement of U-Net structures~\citep{unet} with multi-head self-attention transformer blocks~\cite{transfomer}.

To bridge this gap, we analyze the self-attention mechanism in DiTs by visualizing attention maps between instruction tokens and noised image tokens, as well as among noised image tokens themselves, using two state-of-the-art open-source models involving Omnigen~\cite{omnigen} and FluxEdit~\cite{flux_edit}. Our observations reveal that, after several steps of reverse diffusion, the overall semantics of an instruction, approximated by the average attention weights of all instructions or noised image tokens, tends to focus on the edited regions rather than specific object tokens, distinguishing DiTs from U-Net-based diffusion models. Moreover, different instructions often activate similar attention patterns for a given input image, such as highlighting their respective editing regions and attending to the entire image in the same attention head.

Based on these observations, we propose \textbf{I}nstruction \textbf{I}nfluence \textbf{D}isentanglement (\textbf{IID}), a novel framework for parallel multi-instruction-guided image editing in a single pass, specifically designed for DiT-based models. Concretely, in the initial steps of reverse diffusion, multiple instructions are independently processed. Then, at a designated timestep, we extract attention maps between instruction tokens (noised image tokens) and noised image tokens for FluxEdit (Omnigen) from a predefined layer of the model for each attention head. To disentangle the influence of instruction, we derive instruction-specific masks by comparing head-wise attention maps. For each instruction, we subtract the average attention map of all other instructions from its corresponding attention map in each head, then aggregate results across all heads to generate the final mask. This operation can mitigate interference from the editing regions of other instructions and minimize the influence of non-editing areas. Next, we adaptively concatenate instructions based on their influence scores, estimated by the average attention weight within the editing region, and blend the latent image representations of multiple instructions according to their respective masks. Finally, we construct a new attention mask for the compositional instruction and latent image pair and continue the diffusion process by feeding them into the editing model. To evaluate IID, we conduct quantitative and qualitative experiments on the MagicBrush dataset~\citep{magicbrush} and our collected dataset. Results show that IID outperforms existing baselines in fidelity and instruction completion, demonstrating its effectiveness in parallelized multi-instruction editing.

The contribution of our work is summarized as follows: 1)  We conduct an in-depth investigation on of self-attention mechanisms in DiTs for instruction-guided image editing, uncovering previously unexplored insights that can inform future research. 2) We propose a novel framework that enables the parallel execution of multiple edits in a single denoising process. Our method not only significantly reduces diffusion steps but also improves editing performance, including lower noise generation and better consistency in non-edited regions compared to step-by-step editing. 3) We extensively evaluate our framework on open-source multi-turn editing and custom-constructed datasets, demonstrating its effectiveness.

\section{Related Works}
\subsection{Text-guided editing via Diffusion Models}
Diffusion models have shown remarkable capability in efficiently editing images based on textual conditions. Previous diffusion-based editing methods~\cite{hertz2022prompt,flowedit,MasaCtrl,inversion1,inversionfree1} are built upon text-to-image models and require the caption of the target image and source image as inputs. These approaches employ inversion-based techniques~\cite{ddiminersion,sdedit,inversion2}, where the initial noise map corresponding to the source image is extracted and subsequently denoised to generate the edited image. During this process, the denoising trajectory of the target image can be refined through attention control~\cite{hertz2022prompt,MasaCtrl}, optimization techniques~\cite{inversion1, inversion3} and user-provided masks~\cite{xie2023boxdiff,inversionmask1}, enabling more structurally consistency and semantically coherent modifications. Recently, instruction-guided image editing methods~\cite{hive, geng2024instructdiffusion,smartedit,omniedit, omnigen, shi2024seededit,instructpix2pix} have attracted increasing research interest, as they provide a more user-friendly experience without the need of image captions. These approaches typically fine-tune pretrained text-to-image diffusion models using a conditional image generation objective. Moreover, due to the powerful capabilities of DiTs~\cite{mmDiT,DiT}, recent state-of-the-art editing models (e.g., FluxEdit and OmniGen et al.~\cite{omniedit, omnigen, shi2024seededit}) have shifted their backbone from U-Net-based architectures, such as the Stable Diffusion XL ~\cite{DBLP:conf/iclr/PodellELBDMPR24}, to DiT-based architectures.

\subsection{Multi-instruction Guided image editing}
While instruction-guided image editing models have shown promising results in text-driven editing, they perform well in single-instruction scenarios but struggle with multiple instructions. Simply merging instructions into a single prompt fails to address this limitation due to the dual role of instructions: applying specified modifications while preserving unedited regions. As a result, models often prioritize one instruction over others, leading to instruction conflicts. The most relevant work to our study is FOT~\cite{guo2023focus}, which leverages the cross-attention mechanism of U-Net~\cite{unet} to localize target regions while employing a modulation module to isolate editing areas. However, this approach is not directly applicable to DiT-based frameworks, as effective mask extraction for DiTs remains an open challenge. Moreover, FOT processes all instructions simultaneously, whereas our method handles composite instructions after a predefined step, thus reducing computational overhead. Beyond these, prior works~\cite{huang2025paralleledits, bar2023multidiffusion} integrate multiple diffusion processes via optimization techniques or attention control to address multi-condition image generation in inversion-based editing. However, these methods require task-specific designs and are unsuitable for instruction-guided editing.

\begin{figure*}[h]
\centering
\includegraphics[width=0.9\linewidth]{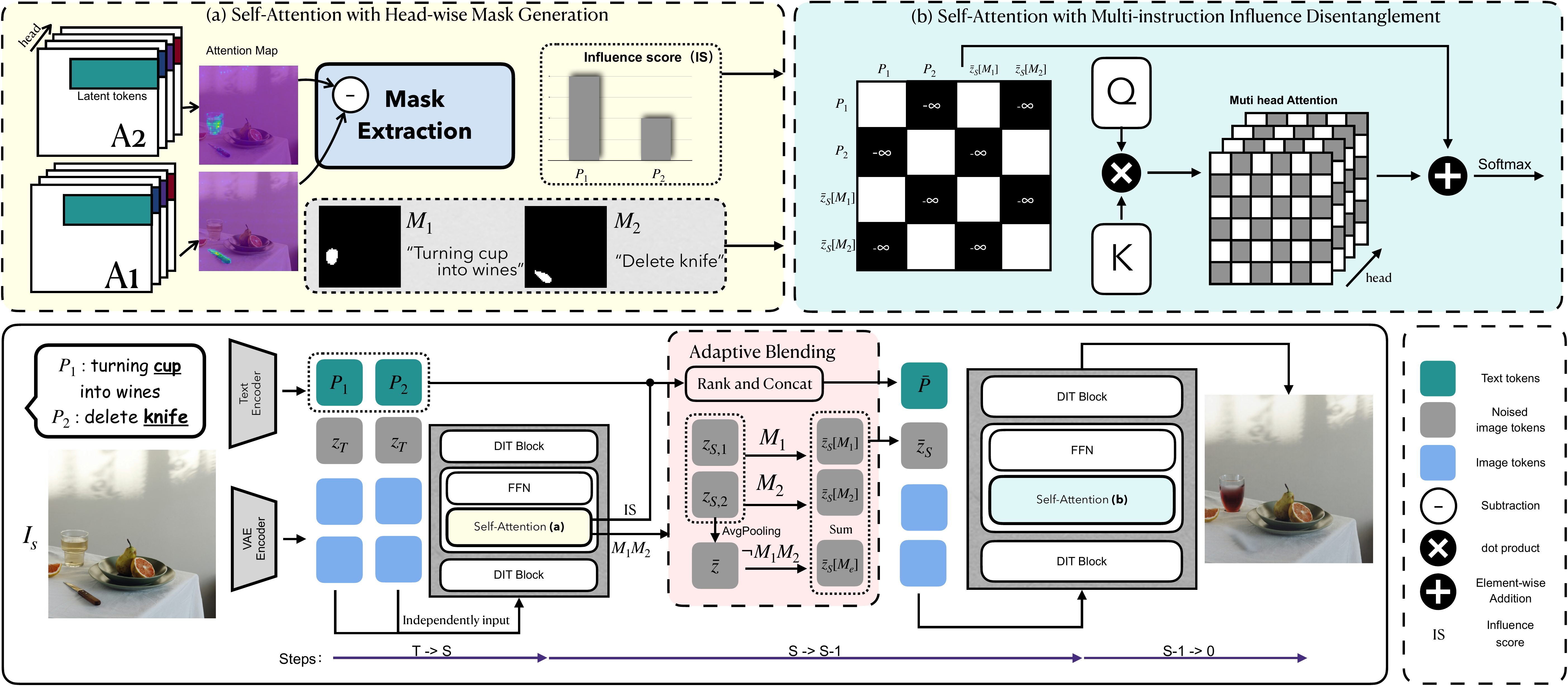}
\caption{Illustration of our proposed IID framework. $T$ denotes the total number of diffusion steps, while $S$ represents the pre-defined step for mask generation and multi-instruction influence disentanglement. $\bar{z}_{S}[M_i]$ corresponds to the token sequence of the noised image $\bar{z}_{S}$ associated with the mask $M_i$ for the instruction $P_i$ (ideally representing the tokens pertinent to the editing area specified by $P_i$).}
\label{fig2}
\vspace{-5pt}
\end{figure*}
\section{Preliminary}
\label{sec:3}
Instruction-guided image editing tasks aim to transform a source image $I_v$ into a target image $I_g$ according to a textual instruction $P$. This section provides an overview of the fundamental concepts of Diffusion Models (DMs) for image editing and self-attention mechanisms in Diffusion Transformers (DiTs).
\subsection{Diffusion Model for Image Editing}
 DMs~\cite{ldm, ddim, ddpm, rectifiedflow} operate through two phases involving a \textit{forward diffusion process} that progressively adds noise to data and a \textit{reverse diffusion process} that aims to reconstruct the original data from noise through iterative denoising of the Gaussian noise. Following Latent Diffusion Models~\cite{ldm}, recent diffusion-based editing models operate in the latent space of a pretrained variational autoencoder (VAE)~\cite{vae} instead of pixel spaces to reduce computational complexity. The source image $I_v $, target image $ I_g$, and text prompt $ P $ are encoded into latent representations as $ c_I $, $z_0 $, and $c_P$ by corresponding encoders. While the forward process corrupts the latent representation $z_0$ of target image by introducing noise $\epsilon$  over $T$ timesteps, the reverse process learns to predict this noise added at each timestep $t$, conditioned on the current noised latent representation $z_t$, time step information $t$, $c_I$ and $c_P$ via a neural network $\epsilon_{\theta}$. Then, the reverse update rule can be expressed as follows:
\begin{equation}
 \label{eq:1}
    z_{t-1} = z_t - \epsilon_{\theta}\left( z_t, t, c_I, c_P, \right),
\end{equation}
where $z_{T}\sim\mathcal{N}(0,1)$ and $\epsilon_{\theta}$ can be obtained by optimizing  the following training objective : 
\begin{equation}
    \mathcal{L} = \mathbb{E}\left[ \left\| \epsilon - \epsilon_{\theta} \left(z_t, t, c_I, c_P \right) \right\|_2^2 \right].
\end{equation}
\subsection{Self-Attention in Diffusion Transformers}
\label{subsec:dit_attention}
To leverage the global receptive and scaling advantages of transformer for image synthesis,
DiTs decompose the noisy input image $z_t$ into  a sequence of $N_z$ patch tokens, denoted as $\{z_{t,i}\}_{i=1}^{N_z}$ where $t$ represent the $t$-th time step. These tokens are subsequently processed through multiple stacked multi-head self-attention layers to predict each timestep's noise $\epsilon$. The fundamental self-attention mechanism is defined as follows:
\begin{equation}
\label{eq:3}
A^j(Q^j,K^j,V^j) = \text{softmax}\left(\frac{Q^j{K^j}^\top}{\sqrt{d}}\right)V^j
\end{equation}
where $Q^j$, $K^j$, and $V^j$ correspond to the query, key, and value matrices, respectively. $d$ represents the dimension scaling factor and $j$ represents the $j$-th attention head.

Although various DiT architectures differ in how they incorporate conditional information, state-of-the-art frameworks such as FluxEdit, using an enhanced version of MM-DiT \cite{mmDiT}, and Omnigen \cite{omnigen} employ heterogeneous token concatenation strategies to enhance the model’s ability to adhere to conditioning signals. Concretely, assume the text instruction $P$ is tokenized into $N_p$ text tokens and encoded as  $\{p_i\}_{i=1}^{N_p}$, the source image $I_v$ is decomposed into $N_v$ patch tokens, and encoded as $\{v_{i}\}_{i=1}^{N_v}$. FluxEdit concatenates the sequence of text tokens and noisy image tokens as input to compute the $Q^j$, $K^j$, and $V^j$ matrices:
\begin{small}
\begin{align}
\label{eq:4}
Q^j &= W_q^j[\{p_{i}\}_{i=1}^{N_p},\{z_{t,i}\}_{i=1}^{N_z}] \notag \\
K^j&= W_k^j[\{p_{i}\}_{i=1}^{N_p},\{z_{t,i}\}_{i=1}^{N_z}] \\
V^j &= W_v^j[\{p_{i}\}_{i=1}^{N_p},\{z_{t,i}\}_{i=1}^{N_z}]\notag, 
\end{align}
\end{small}where $W_q^j$, $W_k^j$ and $W_v^j$ denote projection matrices for $j$-th head. $[~]$ represents a token concatenation operation. Furthermore, FluxEdit incorporates image conditioning $I_v$  with the noisy image before patchifying it into tokens and integrates timestep information through modulation techniques.  
In contrast, OmniGen concatenates all conditional tokens and noisy image tokens to compute the key and value matrices as follows:
\begin{small}
\begin{align}
\label{eq:5}
Q^j &= \begin{cases} 
W_q^j[\{p_{i}\}_{i=1}^{N_p}, \{v_{i}\}_{i=1}^{N_v}, t,\{z_{t,i}\}_{i=1}^{N_z}] & t = T \notag \\
W_q^j[ t,\{z_{t,i}\}_{i=1}^{N_z}] & t < T 
\end{cases} \\
K^j &= W_k^j[\{p_{i}\}_{i=1}^{N_p}, \{v_{i}\}_{i=1}^{N_v}, t,\{z_{t,i}\}_{i=1}^{N_z}] \\
V^j &= W_v^j[\{p_{i}\}_{i=1}^{N_p}, \{v_{i}\}_{i=1}^{N_v}, t,\{z_{t,i}\}_{i=1}^{N_z}]\notag,
\end{align}
\end{small}where at the initial timestep ($t = T$), the query comprises all conditioning and the noisy image tokens. However, for subsequent timesteps ($t < T$), the query is constructed by concatenating the timestep token and noisy image tokens.

\begin{figure*}[h]
\centering
\includegraphics[width=0.8\linewidth]{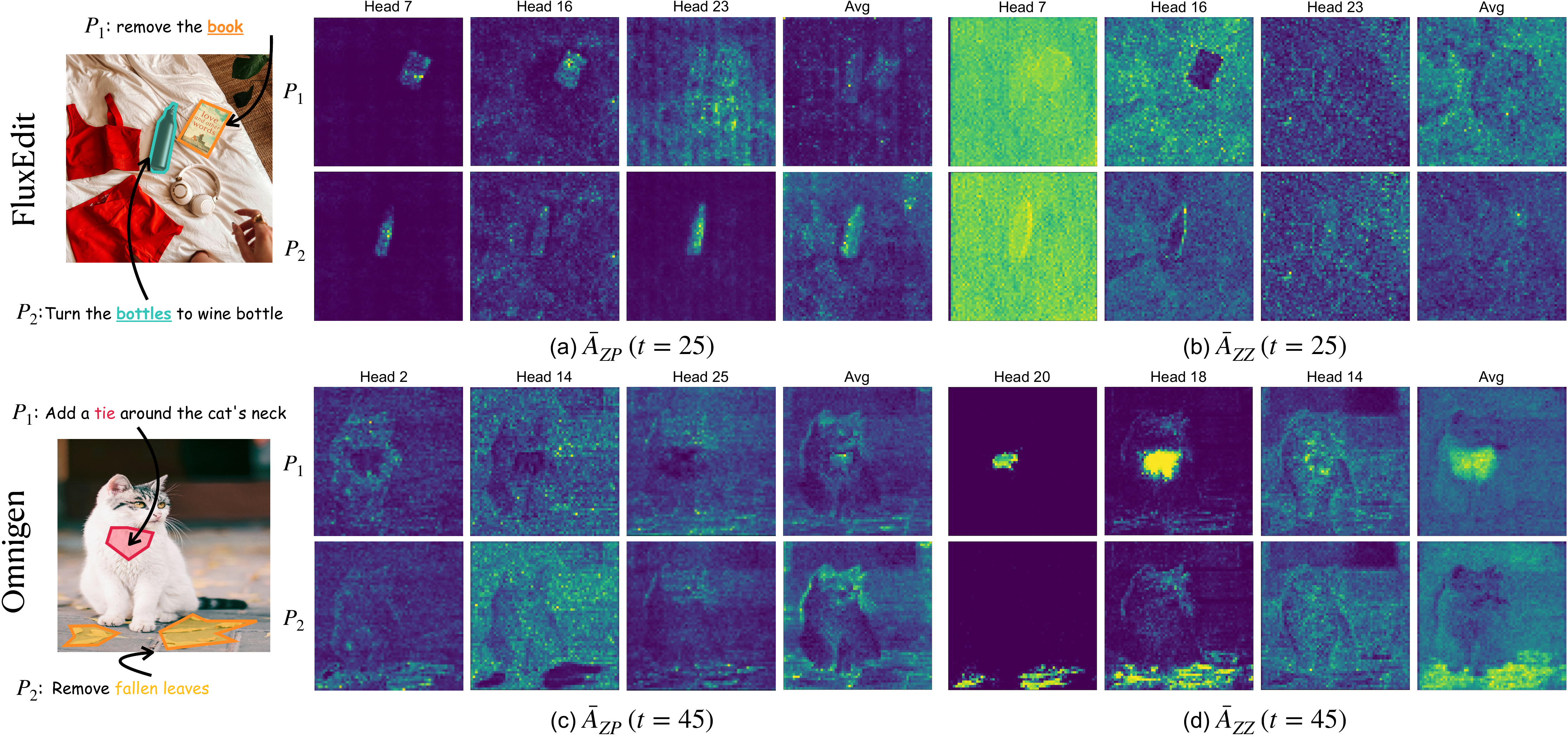}
\caption{The visualization of attention map between the instruction tokens and noise image tokens $\bar{A}_{ZP}$ and among noise image tokens $\bar{A}_{ZZ}$. Attention weights are extracted from the penultimate layer. ``Avg" represents the averaging attention map across all heads.}
\vspace{-5pt}
\label{fig3}
\end{figure*}

\section{Methodology}
In multi-instruction-guided image editing, we are given a set of text instructions $\{P_1, \dots, P_N\}$, where $N$ denotes the number of instructions. Given a source image $I_v$, the objective is to apply all instructions to generate the target image $I_g$ with height $H$ and width $W$. However, when multiple instructions are merged into a single one, the model often prioritizes one edit while neglecting others, resulting in instruction conflicts. To address this, as illustrated in Fig. \ref{fig2}, we first localize the editing region for each instruction using a head-wise mask generation strategy at a predefined step $S$. To enable parallel execution of multiple edits in a single pass, we adaptively concatenate instructions based on their influence scores and blend the latent representations of all instructions based on their respective masks. We then construct a new attention mask to disentangle the influences of various instructions for compositional instruction and latent images. Finally, the model takes these as input to denoise over subsequent timesteps, producing a final image that faithfully completes all intended edits.
\subsection{Attention Weight Analysis}
To mitigate instruction conflicts, it is crucial to ensure that each instruction $P_i$ does not influence regions edited by other instructions through masks. While it remains unexplored whether self-attention in DiTs exhibits similar robust grounding capabilities to prior U-Net-based architectures in image editng tasks, we first analyze the attention weight $A^j$ of the self-attention in DiTs, as defined in Eq. (\ref{eq:3}).


As shown in Eq. (\ref{eq:4}) and Eq. (\ref{eq:5}), FluxEdit and OmniGen employ distinct token concatenation strategies to construct the query, key, and value matrices, resulting in different attention weight distributions. To provide a generalized perspective on DiT architectures, we focus on attention patterns shared across both models. Specifically,  for $j$-th attention head, we analyze the attention weights between the the noisy image token sequence $\{z_i\}_{i=1}^{N_z}$ and instruction token sequence $\{p_i\}_{i=1}^{N_p}$, represented by $A^j_{ZP}\in \mathbb{R}^{N_z \times N_p}$ and the attention weights among the noisy image tokens, denoted as $A^j_{ZZ} \in \mathbb{R}^{N_z \times N_z}$.

Unlike previous U-Net-based methods that extract specific tokens (e.g., those corresponding to edited objects) from attention weights to construct attention maps, we average $A^j_{ZP}$ and $A^j_{ZZ}$ along the second dimension. This approach captures the semantics of the entire instruction, yielding an $N_z$-dimensional vector. The vector is then min-max normalized and reshaped into attention maps as $\bar{A}^j_{ZP}$ and $\bar{A}^j_{ZZ}$, each with a resolution of $H'\times W'$ where $H'=H//q$,  $W'=W//q$ and  $q=16$. Notably, this method eliminates explicit token extraction, enhancing versatility and adaptability across different instruction types.

Then, we visualize both types of attention maps extracted from FluxEdit and Ominigen. As shown in Fig. \ref{fig3}, our key findings are:  1) In FluxEdit, most attention heads in $\bar{A}_{ZP}$ strongly highlight edited regions, demonstrating effective instruction guidance, while a smaller subset distributes attention across the entire image. This pattern emerges after a few diffusion steps, whereas $\bar{A}_{ZZ}$ follows a similar trend but becomes prominent later than $\bar{A}_{ZP}$ (see Appendix. \ref{sec:app1}). 2) Omnigen's $\bar{A}_{ZP}$ does not intensely focus on edited regions; instead, $\bar{A}_{ZZ}$ exhibits apparent attention to these areas even at early timesteps. Given Eq. (\ref{eq:5}), where Omnigen's instruction tokens $\{p_i\}_{i=1}^{N_p}$ only interact with themselves at $t = T$, we suggest that the interaction among textual tokens is insufficient, and textual information gradually propagates into noisy image tokens $\{z_i\}_{i=1}^{N_z}$ over successive timesteps. 3) Despite architectural differences, given the same input image $I_v$ in a multi-instruction setting, many attention heads exhibit similar functionalities across different instructions, such as localizing edits or prioritizing overall image reconstruction. 4) Traditional attention extraction methods based on special tokens and head averaging are suboptimal for DiTs. We speculate that it is possible because self-attention among tokens propagates information holistically rather than relying on individual tokens. 
\subsection{Head-wise Mask Generation}
While previous analyses have shown that each instruction predominantly attends to its corresponding editing region across multiple attention heads, accurately isolating these regions remains challenging. This difficulty arises because certain attention heads maintain a global focus, reducing the contrast between edited and non-edited areas in the averaged attention map. Additionally, even in heads that primarily attend to the editing region, high-intensity noise persists in non-editing areas (e.g., head 16 in Fig. \ref{fig3} (a) and head 18 in Fig. \ref{fig3} (d)), which is difficult to eliminate through post-processing methods such as threshold-based filtering.

Fortunately, we observe that for the same source image, different instructions often exhibit similar attention patterns in the same heads: focusing on the editing region, attending to the global image or exhibiting other preferences. This finding motivates us to design a head-wise mask generation strategy for extracting editing masks. Let’s take the attention map $\bar{A}^j_{ZP}$ between instruction tokens $\{p_i\}_{i=1}^{N_p}$ and noisy image tokens $\{z_i\}_{i=1}^{N_z}$, as an example. For a given instruction $P_i$, we first compute the attention map difference by subtracting the averaging attention map of the same head across all other instructions. The negative values is then set to zero to suppress non-relevant regions. This process can be expressed as follows:
\begin{equation}
M_i^j = \min ( 0, \bar{A}^j_{ZP_i} - \frac{1}{N-1} \sum_{k\neq i}^{N} \bar{A}^j_{ZP_k}),
\end{equation}
where $M_i^j \in \mathbb{R}^{H' \times W'}$ is the editing region-focused attention map for $P_i$ in the $j$-th head.
Such a head-wise subtraction effectively reduces the attention weights in non-editing regions of $M_i^j$, because the attention weights of the same head in these areas are similar across different instructions, leading to near-zero values in $M_i^j$. Moreover, the subtraction also causes the weights corresponding to the editing regions of other instructions to become negative, which are then suppressed to zero in $M_i^j$.

Next, to obtain the final mask $M_i \in \mathbb{R}^{H'\times W'}$ for $P_i$, we first average $M_i^j$ across different heads, then apply a Gaussian Filter to smooth the results, and finally perform binarization using Otsu's Filter~\cite{Otsu}, which automatically determines the threshold without manual intervention. The process can be summarized as follows:
\begin{equation}
M_i = \text{Otsu's Filter}(\text{Gaussian Filter}(\sum_{j=1}^{J}{M_i^j}/J)),
\end{equation}
where $J$ is the number of attention heads.
\subsection{Adaptive Blender}
To enable parallel instruction execution in a signal pass as well as reduce computation, we aim to concatenate all individual instructions into a compositional one, denoted as $\bar{P}$ and aggregate noised images $z_{S, i}$ corresponding to each $P_i$ as a compositional one, denoted as $\bar{z}_{S}$ at the $S$-th timestep. However, the positional information of different instructions plays a crucial role in determining their execution priority during the editing process. For instance, in FluxEdit, earlier instructions tend to be executed more effectively, whereas later ones may not be fully realized. To mitigate this issue, for OmniGen, we ensure all $P_i$ share the same position embedding, thereby neutralizing positional bias. However, for FluxEdit, which does not employ position embeddings, we first compute the influence score of each instruction on its corresponding editing region using  $\sum_{j=1}^{J} (\bar{A}^j_{ZP} \cdot M_i)$ and normalize these score for $P_i$. Then, all instructions can be sorted in ascending order based on these scores to approximately equalize the editing influence across all instructions.

Next, to construct the compositional noisy image $\bar{z}_S$, we blend the latent images $z_{S,i}$ of all instructions using the extracted masks.  First, we compute the averaging latent representation of all instructions at timestep $S$, denoted as $\bar{z}_{S,0}=\sum_{i=1}^Nz_{S,i}/N$. Then, for each instruction, we update the corresponding masked regions in this averaged representation using the respective $z_{S,i}$ as follows:
\begin{equation}
\bar{z}_{S,i} = z_{S,i} \cdot M_i + \bar{z}_{S,i-1}\cdot (1-M_i),
\end{equation}
where $i\in[1,N]$ and $\bar{z}_{S} =\bar{z}_{S,N}$. For overlapping mask regions across multiple instructions, these regions are replaced with the averaged values from the corresponding $z_{S,i}$ to consider the information of all instructions.
\subsection{Multi-instruction Influence Disentanglement}
To disentangle the influence of each instruction to make each one does not interfere with others' editing regions, as illustrated in Fig. \ref{fig2} (b), we construct an attention mask between the tokens of the compositional instruction $\bar{P}$ and $\bar{z}_S$. While instruction tokens attend to all tokens of the noisy image before timestep $S$, we modify this strategy so that the token of $P_i$ can only attend to noisy image tokens excluding regions masked by $M_j$, where $j \in [1, N]$, $j \neq i$. Note that in cases where editing regions overlap, we allow tokens from different instructions to attend to the noisy image tokens in these shared regions. Moreover, in FluxEdit, concatenating instructions introduce interactions among the tokens of various instructions, potentially blurring overall semantics. To mitigate this, we further constrain instructions from attending to each other's tokens. Additionally, for the attention mask among noised image tokens, we let tokens corresponding to each $M_i$ not be seen in other instruction's editing region. Finally, using the newly constructed attention mask, we replace $z_t$ and $C_P$ in Eq. \ref{eq:1} with $z_S$ and the latent representation of $\hat{P}$, respectively, and perform denoising over subsequent timesteps to ensures that the generated image accurately reflects all intended edits.



\section{Experiments}
\subsection{Experiment Setting}
\textbf{Dataset.}
We evaluate the effectiveness of our framework on the test split of the MagicBrush dataset~\cite{magicbrush}, which comprises 535 editing sessions with up to three editing instructions per session. To ensure comparability with existing evaluations, we assess our proposed Instruction Influence Disentanglement (\textbf{IID}) framework across all sessions. For single-instruction samples, we directly the editing model. Additionally, we collect 50 real-world images with varying input sizes and complex instructions to reflect practical editing scenarios for human preference study.

\noindent\textbf{Metrics.} Following \citet{magicbrush}, we utilize L1 and L2 to  quantify pixel-level differences between generated and ground truth images. We adopt CLIP-I and DINO to measure the image quality with the cosine similarity between the generated image and ground truth image using their CLIP~\cite{clip} and DINO~\cite{dino} embeddings. Moreover, we employ CLIP-T~\cite{dreambothtou} to measure the text-image alignment using the cosine similarity between the descriptions of ground truth images and the CLIP embeddings of generated images. 

\noindent\textbf{Baselines.} Due to the absence of multi-instruction-guided image editing approaches specifically designed for DiT-based models, we compare our proposed IID with the step-by-step editing in each session. Our primary evaluation is conducted on two DiT-based instruction-guided editing models: FluxEdit~\cite{flux_edit} and Omnigen~\cite{omnigen}. Since the original FluxEdit underperforms on common editing tasks such as addition, deletion, and modification, we fine-tune FLUX.1-dev\footnote{\url{https://huggingface.co/black-forest-labs/FLUX.1-dev}} it using the same flux control framework as FluxEdit on our private 0.3M high-quality instruction-image pairs. In contrast, Omnigen is a unified image generation model capable of handling complex instruction-guided editing tasks due to its pretraining across diverse computer vision tasks. Additionally, we also consider target caption-guided editing models involving SD-SDEdit~\cite{shi2024seededit} and Null Text Inversion~\cite{inversion1} and other U-Net-based instruction-guided editing models consisting of HIVE~\cite{hive} and InstructPix2Pix~\cite{instructpix2pix}.

\noindent\textbf{Implementation details.} For FluxEdit, we set the pre-defined step as 27, guidance scale to 60 and the total diffusion steps to 30 (i.e., $S=27$ and $N=30$). For Omnigen, we set the pre-defined step as 15, the total diffusion steps to 50 and adopt the default settings of other  perparameters (i.e., $S=15$ and $N=50$). We extract the mask from the penultimate layer of the two models. For other baselines, we use the results reported by \citet{magicbrush}.
\begin{table}
\caption{Quantitative study on diffusion based baselines on the MagicBrush test set. The best results are marked in \textbf{bold}. $\Delta$ denotes the performance improvement achieved by our proposed IID framework compared to the corresponding baseline.}
\label{tab:quantitative_study}
\begin{adjustbox}{width=\linewidth, center}
\centering
\begin{tabular}{cccccc}
\hline
\textbf{Method} & \textbf{L1$\downarrow$} & \textbf{L2$\downarrow$} & \textbf{CLIP-I$\uparrow$} & \textbf{DINO$\uparrow$} & \textbf{CLIP-T$\uparrow$} \\
\hline
\multicolumn{6}{c}{\textit{Target caption-guided}} \\
\hline
SD-SDEdit~\citep{shi2024seededit} & 0.1616 & 0.0602 & 0.7933 & 0.6212 & 0.2694 \\
Null Text Inversion~\cite{inversion1} & 0.1057 & 0.0335 & 0.8468 & 0.7529 & 0.2710 \\
\hline
\multicolumn{6}{c}{\textit{Instruction-guided}} \\
\hline
HIVE~\cite{hive} & 0.1521 & 0.0557 & 0.8004 & 0.6463 & 0.2673 \\
InstructPix2Pix~\cite{instructpix2pix} & 0.1584 & 0.0598 & 0.7924 & 0.6177 & 0.2726 \\
\cdashline{1-6}[2pt/3pt]
Omnigen~\cite{omnigen} & 0.1325 & 0.0543 & 0.8634 & 0.7639 & 0.2820 \\
w/ \textbf{IID}&0.1115 & 0.0466 &0.8714 &0.7902 &\textbf{0.2928} \\
 $\Delta$ (Enhancement) & 0.0210 & 0.0077 & 0.0080 & 0.0263 & 0.0108 \\
\cdashline{1-6}[2pt/3pt]
FluxEdit & 0.1048 & 0.0340 & 0.8487 & 0.7320 & 0.2804\\
w/ \textbf{IID} & \textbf{0.0731} &\textbf{0.0218} & \textbf{0.8855} & \textbf{0.8032} &  0.2837 \\
 $\Delta$ (Enhancement) & 0.0317 & 0.0122 & 0.0368 & 0.0712 & 0.0033 \\
\hline
\end{tabular}
\end{adjustbox}
\end{table}
\begin{table}[h]
\centering
\caption{Human preference study.}
\label{tab:comparison}
\begin{adjustbox}{width=\linewidth, center}
    \begin{tabular}{c|cc|cc|cc}
        \toprule
        \multirow{2}{*}{\textbf{Method}} & 
        \multicolumn{2}{c|}{\textbf{Two-instruction}} & 
        \multicolumn{2}{c|}{\textbf{Three-instruction}} & 
        \multicolumn{2}{c}{\textbf{Four-instruction}} \\
        & \textbf{Instruction} & \textbf{Image} & 
          \textbf{Instruction} & \textbf{Image} & 
          \textbf{Instruction} & \textbf{Image} \\
        & \textbf{Alignment} & \textbf{Alignment} & 
          \textbf{Alignment} & \textbf{Alignment} & 
          \textbf{Alignment} & \textbf{Alignment} \\
        \midrule
        Omnigen &  0 & 0 & 0 & 0.05 & 0 & 0 \\
        Omnigen w/ IID &  0.33& 0.58 & 0.80 & 0.41 & 0.60 & 0.38\\
        \hdashline
        FluxEdit & 0 & 0 & 0 & 0.11 & 0 & 0 \\
        FluxEdit w/ IID & 0.67 & 0.42 & 0.20 & 0.41 & 0.40 & 0.62  \\
        \bottomrule
    \end{tabular}
\end{adjustbox}
\end{table}

\begin{figure*}[h]
\centering
\includegraphics[width=\linewidth]{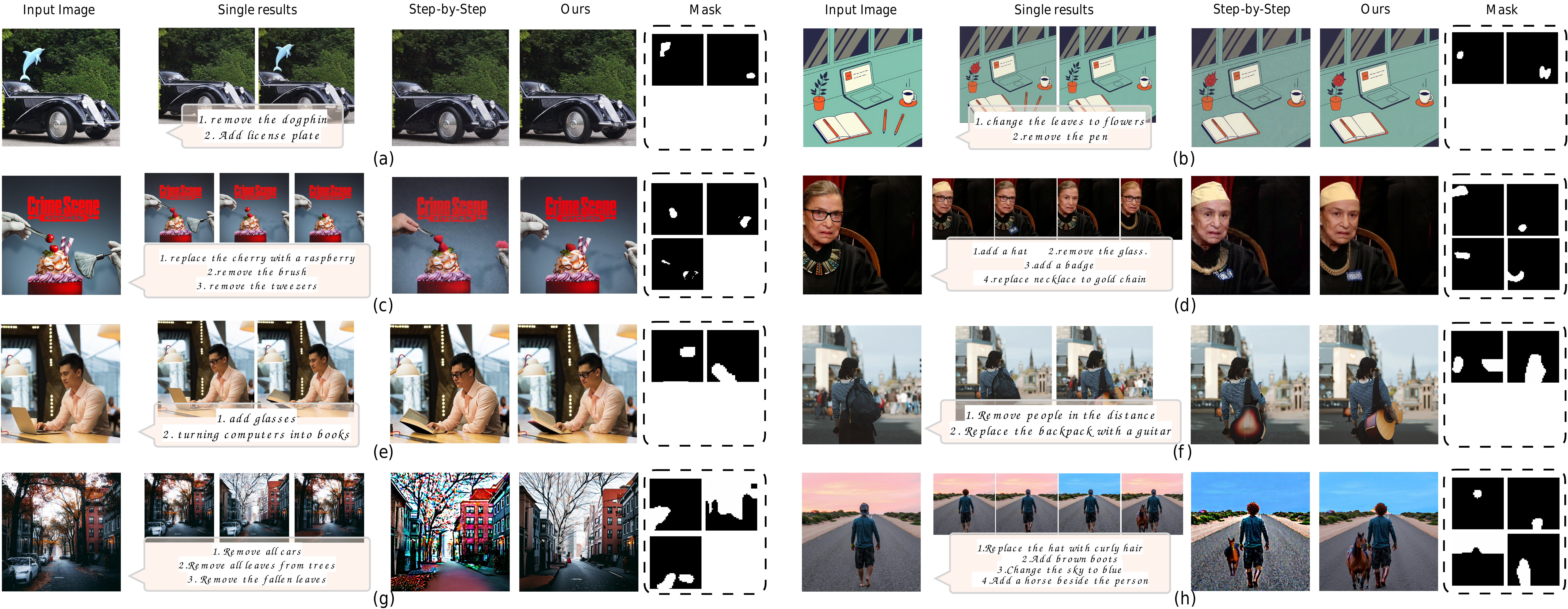}
\caption{Qualitative comparisons. The top two rows of images are based on FluxEdit, while the bottom two rows are based on Omnigen. Single results represent use the one instruction to edit the input image.}
\label{fig4}
\end{figure*}
\subsection{Main Results}
\noindent\textbf{Qualitative evaluation.} We present qualitative experimental results in Fig. \ref{fig4}, demonstrating that our proposed IID framework outperforms the step-by-step approach in two key aspects. First, IID significantly reduces distortion and degradation in the generated images. For instance, as shown in Fig. \ref{fig4} (g), the sequential approach produces images with unrealistic colors and blurry backgrounds, whereas IID preserves image quality by executing instructions in parallel in a single pass, thereby mitigating cumulative errors caused by iterative diffusion processes and VAE encoding-decoding cycles. Second, IID achieves better instruction completion. As seen in Fig. \ref{fig4} (c), our framework successfully completes all three given instructions, whereas the sequential method fails to do so. This can be attributed to IID's ability to generate accurate masks to disentangle the effects of different instructions, ensuring they do not interfere with each other. In contrast, the sequential approach often disrupts the original input image space, resulting in incomplete edits. Moreover, as the number of instructions increases, IID's improvement in editing performance becomes more pronounced. Additionally, all test cases validate the effectiveness of our proposed head-wise mask generation strategy, demonstrating its superiority in capturing editing regions even in complex real-world scenarios without the need to increase mask size.


\noindent\textbf{Quantitative evaluation.}
As shown in Table~\ref{tab:quantitative_study}, the DiT-based models Omnigen and FluxEdit consistently outperform U-Net-based models (HIVE and InstructPix2Pix) across nearly all evaluation metrics. This observation underscores the superiority of DiT architectures as a diffusion modeling framework to generate high-quality images. In particular, FluxEdit achieves the lowest L1 and L2 distances, indicating minimal pixel-level discrepancies from the ground truth, while Omnigen achieves the highest CLIP-I and DINO scores, highlighting its strong capability in accurate instruction execution and perceptual alignment .Furthermore, our proposed IID framework significantly improves performance over the step-by-step instruction editing approach. Specifically, IID enhances both Omnigen and FluxEdit across all key metrics, with notable reductions in L1/L2 errors and increases in CLIP-I, DINO, and CLIP-T scores. These improvements stem from IID’s ability to mitigate degradation caused by iterative denoising and repeated encoding-decoding through VAE, which can otherwise introduce noise and artifacts. By disentangling multi-instruction influences during the editing process, IID ensures better semantic consistency and structural preservation, leading to sharper, more coherent image edits.

\noindent\textbf{Human Preference Study.} 
Following \citet{guo2023focus}, we conduct a human preference study using 50 real-world multi-run editing scenarios, involving 5 participants. For instruction alignment, participants are asked to select the method that best matches the editing effect specified by the instruction. For image alignment, they are asked to choose the method that best preserves the original image details. As shown in Table \ref{tab:comparison}, our framework demonstrates notable superiority compared to step-by-step editing in terms of both instruction completion and the consistency between the pre-edit and post-edit images. Additionally, in complex instruction scenarios, OmniGen performs well in understanding and executing instructions, but it is more prone to generating distortions as the number of instructions increases.
\subsection{Ablation Study}
In the Appendix, we conduct extensive ablation studies to analyze key design choices in our framework. First, we examine the evolution of attention maps across different attention heads in the penultimate layer of FluxEdit and OmniGen in the Appendix. \ref{sec:app1}. The results reveal that for FluxEdit both $\bar{A}_{ZP}$ and $\bar{A}_{ZZ}$ gradually refine their focus on editing regions as diffusion progresses, while only $\bar{A}_{ZZ}$ show such trend for  Omnigen. As such, for FluxEdit, both $\bar{A}_{ZP}$ and $\bar{A}_{ZZ}$ can be used to generate masks while only $\bar{A}_{ZZ}$ is suitable for Omnigen. Second, we evaluated the impact of different timesteps and extracted layers on mask generation quality in the Appendix. \ref{sec:app2}. The results show that the progression of diffusion steps significantly affects mask quality, with improvements observed as the steps increase. However, after a specific number of steps, the masks stabilize, suggesting that an optimal range of timesteps exists for achieving high-quality masks. Moreover, different model layers contribute varying levels of semantic and spatial focus for mask generation. Lower layers fail to generate focused masks due to dispersed attention weights, while the final layer prioritizes image reconstruction over mask quality. The penultimate layer strikes the right balance, providing semantically rich and spatially focused masks, making it the most suitable layer for mask extraction. Lastly, we explore the influence of the predefined step $S$ on the editing performance in Appendix \ref{sec:app3}. The results show that the choice of step $S$ is critical for OmniGen's ability to balance instruction fulfillment and image consistency. When $S$ is too large (e.g., $S \geq 45$), the generated images fail to adhere to the editing instructions and deviate significantly from the original image, while choosing a smaller step, such as $S = 15$ can avoid the disruption of instruction-related semantic information during latent blending. However, FluxEdit demonstrates resilience in maintaining the original feature space during early reverse diffusion steps.


\section{Conclusion}
In this work, we conduct an in-depth investigation of self-attention mechanisms in DiTs for instruction-guided image editing, uncovering key insights that could inform future research. Based on these findings, we introduce IID, a novel framework that enables parallel execution of multiple edits in a single denoising process, reducing diffusion steps while improving editing quality compared to step-by-step editing. Our approach first localizes editing regions using a head-wise mask generation strategy at a predefined step. To achieve simultaneous editing, we adaptively concatenate instructions based on their influence scores and blend latent representations of multiple instructions to construct the noisy image input for the composite instruction. Finally, we construct an attention mask to mitigate instruction conflicts and let the editing model take these as input to denoise over subsequent timesteps. Extensive evaluations on an open-source multi-turn editing dataset and custom benchmarks demonstrate the effectiveness of our method. {
    \small
    \bibliographystyle{ieeenat_fullname}
    \bibliography{main}

\begin{thebibliography}{45}
\providecommand{\natexlab}[1]{#1}
\providecommand{\url}[1]{\texttt{#1}}
\expandafter\ifx\csname urlstyle\endcsname\relax
  \providecommand{\doi}[1]{doi: #1}\else
  \providecommand{\doi}{doi: \begingroup \urlstyle{rm}\Url}\fi

\bibitem[Avrahami et~al.(2023)Avrahami, Fried, and Lischinski]{inversionmask1}
Omri Avrahami, Ohad Fried, and Dani Lischinski.
\newblock Blended latent diffusion.
\newblock \emph{{ACM} Trans. Graph.}, 42\penalty0 (4):\penalty0 149:1--149:11, 2023.

\bibitem[Avrahami et~al.(2024)Avrahami, Patashnik, Fried, Nemchinov, Aberman, Lischinski, and Cohen{-}Or]{inversion2}
Omri Avrahami, Or Patashnik, Ohad Fried, Egor Nemchinov, Kfir Aberman, Dani Lischinski, and Daniel Cohen{-}Or.
\newblock Stable flow: Vital layers for training-free image editing.
\newblock \emph{CoRR}, abs/2411.14430, 2024.

\bibitem[Bar-Tal et~al.(2023)Bar-Tal, Yariv, Lipman, and Dekel]{bar2023multidiffusion}
Omer Bar-Tal, Lior Yariv, Yaron Lipman, and Tali Dekel.
\newblock Multidiffusion: Fusing diffusion paths for controlled image generation.
\newblock \emph{Proceedings of Machine Learning Research}, 202:\penalty0 1737--1752, 2023.

\bibitem[Brooks et~al.(2023)Brooks, Holynski, and Efros]{instructpix2pix}
Tim Brooks, Aleksander Holynski, and Alexei~A Efros.
\newblock Instructpix2pix: Learning to follow image editing instructions.
\newblock In \emph{Proceedings of the IEEE/CVF conference on computer vision and pattern recognition}, pages 18392--18402, 2023.

\bibitem[Cao et~al.(2023)Cao, Wang, Qi, Shan, Qie, and Zheng]{MasaCtrl}
Mingdeng Cao, Xintao Wang, Zhongang Qi, Ying Shan, Xiaohu Qie, and Yinqiang Zheng.
\newblock Masactrl: Tuning-free mutual self-attention control for consistent image synthesis and editing.
\newblock In \emph{{IEEE/CVF} International Conference on Computer Vision, {ICCV} 2023, Paris, France, October 1-6, 2023}, pages 22503--22513. {IEEE}, 2023.

\bibitem[Caron et~al.(2021)Caron, Touvron, Misra, J{\'{e}}gou, Mairal, Bojanowski, and Joulin]{dino}
Mathilde Caron, Hugo Touvron, Ishan Misra, Herv{\'{e}} J{\'{e}}gou, Julien Mairal, Piotr Bojanowski, and Armand Joulin.
\newblock Emerging properties in self-supervised vision transformers.
\newblock In \emph{2021 {IEEE/CVF} International Conference on Computer Vision, {ICCV} 2021, Montreal, QC, Canada, October 10-17, 2021}, pages 9630--9640. {IEEE}, 2021.

\bibitem[Esser et~al.(2024)Esser, Kulal, Blattmann, Entezari, M{\"u}ller, Saini, Levi, Lorenz, Sauer, Boesel, et~al.]{mmDiT}
Patrick Esser, Sumith Kulal, Andreas Blattmann, Rahim Entezari, Jonas M{\"u}ller, Harry Saini, Yam Levi, Dominik Lorenz, Axel Sauer, Frederic Boesel, et~al.
\newblock Scaling rectified flow transformers for high-resolution image synthesis.
\newblock In \emph{Forty-first international conference on machine learning}, 2024.

\bibitem[Geng et~al.(2024)Geng, Yang, Hang, Li, Gu, Zhang, Bao, Zhang, Li, Hu, et~al.]{geng2024instructdiffusion}
Zigang Geng, Binxin Yang, Tiankai Hang, Chen Li, Shuyang Gu, Ting Zhang, Jianmin Bao, Zheng Zhang, Houqiang Li, Han Hu, et~al.
\newblock Instructdiffusion: A generalist modeling interface for vision tasks.
\newblock In \emph{Proceedings of the IEEE/CVF Conference on computer vision and pattern recognition}, pages 12709--12720, 2024.

\bibitem[Gu et~al.(2024)Gu, Li, Zhang, Chen, Wen, Luo, and Zhu]{mutlireward}
Xin Gu, Ming Li, Libo Zhang, Fan Chen, Longyin Wen, Tiejian Luo, and Sijie Zhu.
\newblock Multi-reward as condition for instruction-based image editing.
\newblock \emph{CoRR}, abs/2411.04713, 2024.

\bibitem[Guo and Lin(2024)]{guo2023focus}
Qin Guo and Tianwei Lin.
\newblock Focus on your instruction: Fine-grained and multi-instruction image editing by attention modulation.
\newblock In \emph{Proceedings of the IEEE/CVF Conference on Computer Vision and Pattern Recognition (CVPR)}, 2024.

\bibitem[Hertz et~al.(2022)Hertz, Mokady, Tenenbaum, Aberman, Pritch, and Cohen-Or]{hertz2022prompt}
Amir Hertz, Ron Mokady, Jay Tenenbaum, Kfir Aberman, Yael Pritch, and Daniel Cohen-Or.
\newblock Prompt-to-prompt image editing with cross attention control.
\newblock \emph{arXiv preprint arXiv:2208.01626}, 2022.

\bibitem[Ho et~al.(2020)Ho, Jain, and Abbeel]{ddpm}
Jonathan Ho, Ajay Jain, and Pieter Abbeel.
\newblock Denoising diffusion probabilistic models.
\newblock \emph{Advances in neural information processing systems}, 33:\penalty0 6840--6851, 2020.

\bibitem[Huang et~al.(2025)Huang, Cai, Jia, Lokhande, and Lyu]{huang2025paralleledits}
Mingzhen Huang, Jialing Cai, Shan Jia, Vishnu Lokhande, and Siwei Lyu.
\newblock Paralleledits: Efficient multi-aspect text-driven image editing with attention grouping.
\newblock \emph{Advances in Neural Information Processing Systems}, 37:\penalty0 22569--22595, 2025.

\bibitem[Huang et~al.(2024)Huang, Xie, Wang, Yuan, Cun, Ge, Zhou, Dong, Huang, Zhang, et~al.]{smartedit}
Yuzhou Huang, Liangbin Xie, Xintao Wang, Ziyang Yuan, Xiaodong Cun, Yixiao Ge, Jiantao Zhou, Chao Dong, Rui Huang, Ruimao Zhang, et~al.
\newblock Smartedit: Exploring complex instruction-based image editing with multimodal large language models.
\newblock In \emph{Proceedings of the IEEE/CVF Conference on Computer Vision and Pattern Recognition}, pages 8362--8371, 2024.

\bibitem[Jawahar et~al.(2019)Jawahar, Sagot, and Seddah]{DBLP:conf/acl/JawaharSS19}
Ganesh Jawahar, Beno{\^{\i}}t Sagot, and Djam{\'{e}} Seddah.
\newblock What does {BERT} learn about the structure of language?
\newblock In \emph{Proceedings of the 57th Conference of the Association for Computational Linguistics, {ACL} 2019, Florence, Italy, July 28- August 2, 2019, Volume 1: Long Papers}, pages 3651--3657. Association for Computational Linguistics, 2019.

\bibitem[Joseph et~al.(2024)Joseph, Udhayanan, Shukla, Agarwal, Karanam, Goswami, and Srinivasan]{seq1}
K.~J. Joseph, Prateksha Udhayanan, Tripti Shukla, Aishwarya Agarwal, Srikrishna Karanam, Koustava Goswami, and Balaji~Vasan Srinivasan.
\newblock Iterative multi-granular image editing using diffusion models.
\newblock In \emph{{IEEE/CVF} Winter Conference on Applications of Computer Vision, {WACV} 2024, Waikoloa, HI, USA, January 3-8, 2024}, pages 8092--8101. {IEEE}, 2024.

\bibitem[Kawar et~al.(2023)Kawar, Zada, Lang, Tov, Chang, Dekel, Mosseri, and Irani]{inversion3}
Bahjat Kawar, Shiran Zada, Oran Lang, Omer Tov, Huiwen Chang, Tali Dekel, Inbar Mosseri, and Michal Irani.
\newblock Imagic: Text-based real image editing with diffusion models.
\newblock In \emph{{IEEE/CVF} Conference on Computer Vision and Pattern Recognition, {CVPR} 2023, Vancouver, BC, Canada, June 17-24, 2023}, pages 6007--6017. {IEEE}, 2023.

\bibitem[Khodadadeh et~al.(2022)Khodadadeh, Ghadar, Motiian, Lin, B{\"{o}}l{\"{o}}ni, and Kalarot]{multi1}
Siavash Khodadadeh, Shabnam Ghadar, Saeid Motiian, Wei{-}An Lin, Ladislau B{\"{o}}l{\"{o}}ni, and Ratheesh Kalarot.
\newblock Latent to latent: {A} learned mapper for identity preserving editing of multiple face attributes in stylegan-generated images.
\newblock In \emph{{IEEE/CVF} Winter Conference on Applications of Computer Vision, {WACV} 2022, Waikoloa, HI, USA, January 3-8, 2022}, pages 3677--3685. {IEEE}, 2022.

\bibitem[Kingma and Welling(2014)]{vae}
Diederik~P. Kingma and Max Welling.
\newblock Auto-encoding variational bayes.
\newblock In \emph{2nd International Conference on Learning Representations, {ICLR} 2014, Banff, AB, Canada, April 14-16, 2014, Conference Track Proceedings}, 2014.

\bibitem[Kulikov et~al.(2024)Kulikov, Kleiner, Huberman{-}Spiegelglas, and Michaeli]{flowedit}
Vladimir Kulikov, Matan Kleiner, Inbar Huberman{-}Spiegelglas, and Tomer Michaeli.
\newblock Flowedit: Inversion-free text-based editing using pre-trained flow models.
\newblock \emph{CoRR}, abs/2412.08629, 2024.

\bibitem[Lin et~al.(2024)Lin, Chen, Wang, An, Wang, Tian, Liu, Dai, Wang, and Wang]{schedule}
Haonan Lin, Yan Chen, Jiahao Wang, Wenbin An, Mengmeng Wang, Feng Tian, Yong Liu, Guang Dai, Jingdong Wang, and Qianying Wang.
\newblock Schedule your edit: {A} simple yet effective diffusion noise schedule for image editing.
\newblock In \emph{Advances in Neural Information Processing Systems 38: Annual Conference on Neural Information Processing Systems 2024, NeurIPS 2024, Vancouver, BC, Canada, December 10 - 15, 2024}, 2024.

\bibitem[Liu et~al.(2024)Liu, Wang, Sun, Tian, Kong, Dong, and Li]{DBLP:journals/corr/abs-2406-11890}
Hui Liu, Wenya Wang, Hao Sun, Chris~Xing Tian, Chenqi Kong, Xin Dong, and Haoliang Li.
\newblock Unraveling the mechanics of learning-based demonstration selection for in-context learning.
\newblock \emph{CoRR}, abs/2406.11890, 2024.

\bibitem[Liu et~al.(2022)Liu, Gong, and Liu]{rectifiedflow}
Xingchao Liu, Chengyue Gong, and Qiang Liu.
\newblock Flow straight and fast: Learning to generate and transfer data with rectified flow.
\newblock \emph{arXiv preprint arXiv:2209.03003}, 2022.

\bibitem[Ma et~al.(2024)Ma, Ji, Ye, Lin, Wang, Zheng, Zhou, Sun, and Ji]{I2EBench}
Yiwei Ma, Jiayi Ji, Ke Ye, Weihuang Lin, Zhibin Wang, Yonghan Zheng, Qiang Zhou, Xiaoshuai Sun, and Rongrong Ji.
\newblock I2ebench: {A} comprehensive benchmark for instruction-based image editing.
\newblock In \emph{Advances in Neural Information Processing Systems 38: Annual Conference on Neural Information Processing Systems 2024, NeurIPS 2024, Vancouver, BC, Canada, December 10 - 15, 2024}, 2024.

\bibitem[Meng et~al.(2022)Meng, He, Song, Song, Wu, Zhu, and Ermon]{sdedit}
Chenlin Meng, Yutong He, Yang Song, Jiaming Song, Jiajun Wu, Jun{-}Yan Zhu, and Stefano Ermon.
\newblock Sdedit: Guided image synthesis and editing with stochastic differential equations.
\newblock In \emph{The Tenth International Conference on Learning Representations, {ICLR} 2022, Virtual Event, April 25-29, 2022}. OpenReview.net, 2022.

\bibitem[Mokady et~al.(2023)Mokady, Hertz, Aberman, Pritch, and Cohen{-}Or]{inversion1}
Ron Mokady, Amir Hertz, Kfir Aberman, Yael Pritch, and Daniel Cohen{-}Or.
\newblock Null-text inversion for editing real images using guided diffusion models.
\newblock In \emph{{IEEE/CVF} Conference on Computer Vision and Pattern Recognition, {CVPR} 2023, Vancouver, BC, Canada, June 17-24, 2023}, pages 6038--6047. {IEEE}, 2023.

\bibitem[Otsu(1979)]{Otsu}
Nobuyuki Otsu.
\newblock A threshold selection method from gray-level histograms.
\newblock \emph{{IEEE} Trans. Syst. Man Cybern.}, 9\penalty0 (1):\penalty0 62--66, 1979.

\bibitem[Paul(2024)]{flux_edit}
Sayak Paul.
\newblock Flux.1-dev-edit-v0.
\newblock \url{https://huggingface.co/sayakpaul/FLUX.1-dev-edit-v0}, 2024.
\newblock Accessed: March 7, 2025.

\bibitem[Peebles and Xie(2023)]{DiT}
William Peebles and Saining Xie.
\newblock Scalable diffusion models with transformers.
\newblock In \emph{Proceedings of the IEEE/CVF international conference on computer vision}, pages 4195--4205, 2023.

\bibitem[Podell et~al.(2024)Podell, English, Lacey, Blattmann, Dockhorn, M{\"{u}}ller, Penna, and Rombach]{DBLP:conf/iclr/PodellELBDMPR24}
Dustin Podell, Zion English, Kyle Lacey, Andreas Blattmann, Tim Dockhorn, Jonas M{\"{u}}ller, Joe Penna, and Robin Rombach.
\newblock {SDXL:} improving latent diffusion models for high-resolution image synthesis.
\newblock In \emph{The Twelfth International Conference on Learning Representations, {ICLR} 2024, Vienna, Austria, May 7-11, 2024}. OpenReview.net, 2024.

\bibitem[Radford et~al.(2021)Radford, Kim, Hallacy, Ramesh, Goh, Agarwal, Sastry, Askell, Mishkin, Clark, Krueger, and Sutskever]{clip}
Alec Radford, Jong~Wook Kim, Chris Hallacy, Aditya Ramesh, Gabriel Goh, Sandhini Agarwal, Girish Sastry, Amanda Askell, Pamela Mishkin, Jack Clark, Gretchen Krueger, and Ilya Sutskever.
\newblock Learning transferable visual models from natural language supervision.
\newblock In \emph{Proceedings of the 38th International Conference on Machine Learning, {ICML} 2021, 18-24 July 2021, Virtual Event}, pages 8748--8763. {PMLR}, 2021.

\bibitem[Rombach et~al.(2022)Rombach, Blattmann, Lorenz, Esser, and Ommer]{ldm}
Robin Rombach, Andreas Blattmann, Dominik Lorenz, Patrick Esser, and Bj{\"{o}}rn Ommer.
\newblock High-resolution image synthesis with latent diffusion models.
\newblock In \emph{{IEEE/CVF} Conference on Computer Vision and Pattern Recognition, {CVPR} 2022, New Orleans, LA, USA, June 18-24, 2022}, pages 10674--10685. {IEEE}, 2022.

\bibitem[Ronneberger et~al.(2015)Ronneberger, Fischer, and Brox]{unet}
Olaf Ronneberger, Philipp Fischer, and Thomas Brox.
\newblock U-net: Convolutional networks for biomedical image segmentation.
\newblock In \emph{Medical Image Computing and Computer-Assisted Intervention - {MICCAI} 2015 - 18th International Conference Munich, Germany, October 5 - 9, 2015, Proceedings, Part {III}}, pages 234--241. Springer, 2015.

\bibitem[Ruiz et~al.(2023)Ruiz, Li, Jampani, Pritch, Rubinstein, and Aberman]{dreambothtou}
Nataniel Ruiz, Yuanzhen Li, Varun Jampani, Yael Pritch, Michael Rubinstein, and Kfir Aberman.
\newblock Dreambooth: Fine tuning text-to-image diffusion models for subject-driven generation.
\newblock In \emph{{IEEE/CVF} Conference on Computer Vision and Pattern Recognition, {CVPR} 2023, Vancouver, BC, Canada, June 17-24, 2023}, pages 22500--22510. {IEEE}, 2023.

\bibitem[Sheynin et~al.(2024)Sheynin, Polyak, Singer, Kirstain, Zohar, Ashual, Parikh, and Taigman]{emudit}
Shelly Sheynin, Adam Polyak, Uriel Singer, Yuval Kirstain, Amit Zohar, Oron Ashual, Devi Parikh, and Yaniv Taigman.
\newblock Emu edit: Precise image editing via recognition and generation tasks.
\newblock In \emph{{IEEE/CVF} Conference on Computer Vision and Pattern Recognition, {CVPR} 2024, Seattle, WA, USA, June 16-22, 2024}, pages 8871--8879. {IEEE}, 2024.

\bibitem[Shi et~al.(2024)Shi, Wang, and Huang]{shi2024seededit}
Yichun Shi, Peng Wang, and Weilin Huang.
\newblock Seededit: Align image re-generation to image editing.
\newblock \emph{arXiv preprint arXiv:2411.06686}, 2024.

\bibitem[Song et~al.(2020)Song, Meng, and Ermon]{ddim}
Jiaming Song, Chenlin Meng, and Stefano Ermon.
\newblock Denoising diffusion implicit models.
\newblock \emph{arXiv preprint arXiv:2010.02502}, 2020.

\bibitem[Song et~al.(2021)Song, Meng, and Ermon]{ddiminersion}
Jiaming Song, Chenlin Meng, and Stefano Ermon.
\newblock Denoising diffusion implicit models.
\newblock In \emph{9th International Conference on Learning Representations, {ICLR} 2021, Virtual Event, Austria, May 3-7, 2021}. OpenReview.net, 2021.

\bibitem[Vaswani et~al.(2017)Vaswani, Shazeer, Parmar, Uszkoreit, Jones, Gomez, Kaiser, and Polosukhin]{transfomer}
Ashish Vaswani, Noam Shazeer, Niki Parmar, Jakob Uszkoreit, Llion Jones, Aidan~N. Gomez, Lukasz Kaiser, and Illia Polosukhin.
\newblock Attention is all you need.
\newblock In \emph{Advances in Neural Information Processing Systems 30: Annual Conference on Neural Information Processing Systems 2017, December 4-9, 2017, Long Beach, CA, {USA}}, pages 5998--6008, 2017.

\bibitem[Wei et~al.(2024)Wei, Xiong, Ren, Du, Zhang, and Chen]{omniedit}
Cong Wei, Zheyang Xiong, Weiming Ren, Xeron Du, Ge Zhang, and Wenhu Chen.
\newblock Omniedit: Building image editing generalist models through specialist supervision.
\newblock In \emph{The Thirteenth International Conference on Learning Representations}, 2024.

\bibitem[Xiao et~al.(2024)Xiao, Wang, Zhou, Yuan, Xing, Yan, Wang, Huang, and Liu]{omnigen}
Shitao Xiao, Yueze Wang, Junjie Zhou, Huaying Yuan, Xingrun Xing, Ruiran Yan, Shuting Wang, Tiejun Huang, and Zheng Liu.
\newblock Omnigen: Unified image generation.
\newblock \emph{arXiv preprint arXiv:2409.11340}, 2024.

\bibitem[Xie et~al.(2023)Xie, Li, Huang, Liu, Zhang, Zheng, and Shou]{xie2023boxdiff}
Jinheng Xie, Yuexiang Li, Yawen Huang, Haozhe Liu, Wentian Zhang, Yefeng Zheng, and Mike~Zheng Shou.
\newblock Boxdiff: Text-to-image synthesis with training-free box-constrained diffusion.
\newblock In \emph{Proceedings of the IEEE/CVF International Conference on Computer Vision}, pages 7452--7461, 2023.

\bibitem[Xu et~al.(2023)Xu, Huang, Pan, Ma, and Chai]{inversionfree1}
Sihan Xu, Yidong Huang, Jiayi Pan, Ziqiao Ma, and Joyce Chai.
\newblock Inversion-free image editing with natural language.
\newblock \emph{CoRR}, abs/2312.04965, 2023.

\bibitem[Zhang et~al.(2023)Zhang, Mo, Chen, Sun, and Su]{magicbrush}
Kai Zhang, Lingbo Mo, Wenhu Chen, Huan Sun, and Yu Su.
\newblock Magicbrush: {A} manually annotated dataset for instruction-guided image editing.
\newblock In \emph{Advances in Neural Information Processing Systems 36: Annual Conference on Neural Information Processing Systems 2023, NeurIPS 2023, New Orleans, LA, USA, December 10 - 16, 2023}, 2023.

\bibitem[Zhang et~al.(2024)Zhang, Yang, Feng, Qin, Chen, Yu, Chen, Wang, Savarese, Ermon, et~al.]{hive}
Shu Zhang, Xinyi Yang, Yihao Feng, Can Qin, Chia-Chih Chen, Ning Yu, Zeyuan Chen, Huan Wang, Silvio Savarese, Stefano Ermon, et~al.
\newblock Hive: Harnessing human feedback for instructional visual editing.
\newblock In \emph{Proceedings of the IEEE/CVF Conference on Computer Vision and Pattern Recognition}, pages 9026--9036, 2024.

\end{thebibliography}
}
\clearpage
\setcounter{page}{1}
\maketitlesupplementary
\setcounter{section}{0}
\renewcommand\thesection{\Alph{section}}
\section{Further Analysis of Attention Map}
\label{sec:app1}
\begin{figure*}[h]
\centering
\includegraphics[width=0.9\linewidth]{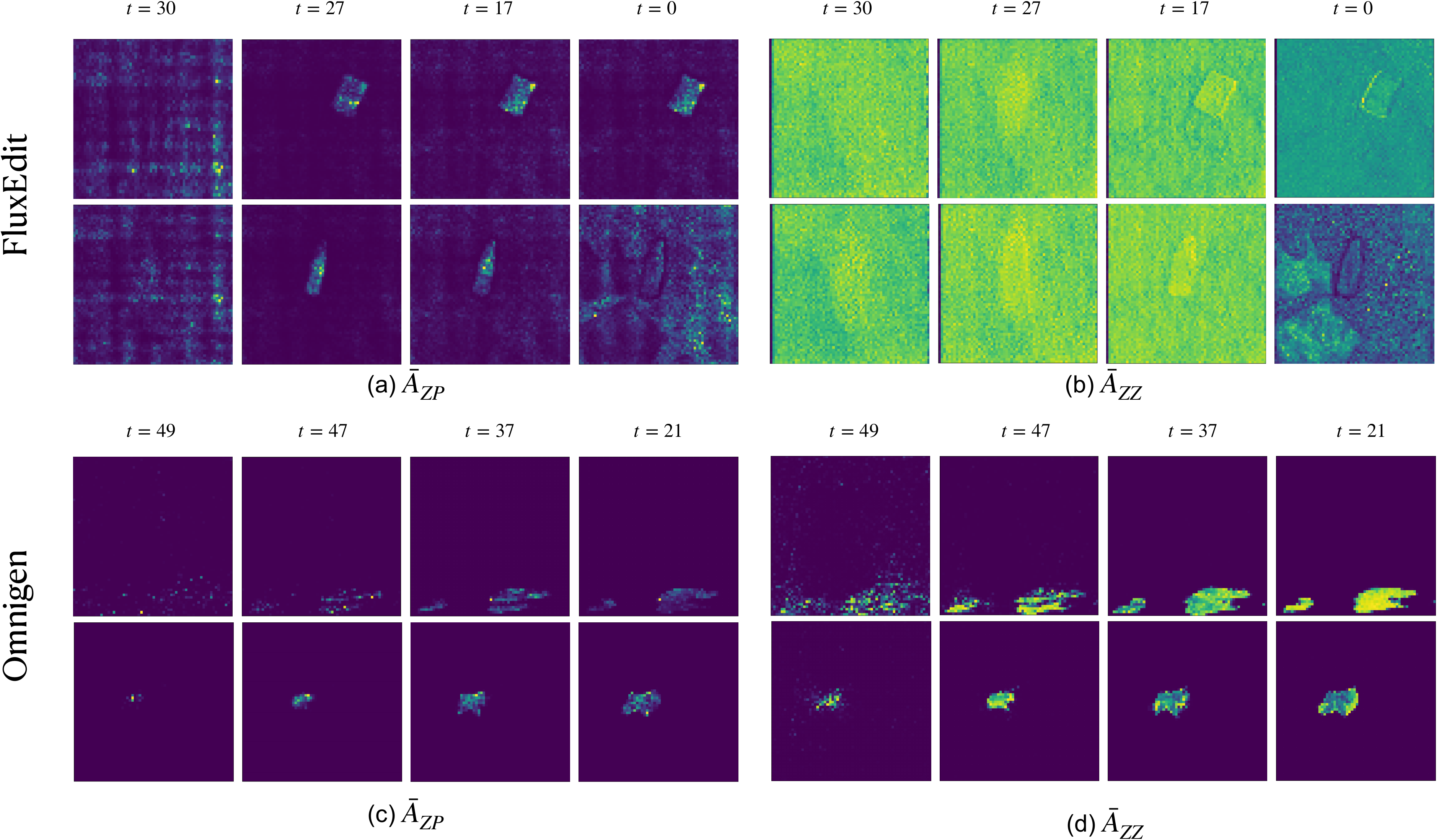}
\caption{Ablation study on the influence of timesteps on the attention maps of the penultimate layer of both models. For $A_{ZP}$ of  Omnigen, we choose the attention head with highest activation to the editing region for displaying.}
\label{fig5}
\end{figure*}

In this section, we analyze the evolution of attention maps across different attention heads in the penultimate layer of FluxEdit and Omnigen, focusing on the attention weights assigned to edited regions. As shown in Fig. \ref{fig5}, for $\bar{A}_{ZP}$ of FluxEdit, at initial timesteps ($t = 30$), $\bar{A}_{ZP}$ exhibits a striped pattern and does not focus on the edited regions. As diffusion progresses, instruction tokens increasingly attend to the editing regions, with this effect becoming more pronounced over successive steps. However, as the noised image approaches the true data distribution ($t \to 0$), the contours of the editing regions in $\bar{A}_{ZP}$ become less distinct than in earlier diffusion stages. In comparison, $\bar{A}_{ZZ}$ begins to focus on the editing regions at later timesteps but eventually follows a similar trend to $\bar{A}_{ZP}$ during the final stages of diffusion.
For $\bar{A}_{ZP}$ of Omnigen, even when visualizing the attention head with the highest activation for the editing region (which represents only a minority among all heads), the attention weights relevant to the edited areas remain insufficiently distinct. This results in some editing regions being overlooked when generating the final mask. Instead, $\bar{A}_{ZZ}$ increasingly focuses on the edited regions as the diffusion steps progress. Moreover, with more diffusion steps, the focus areas for certain instructions slightly expand, which may lead to masks that include regions unrelated to the instruction. Thus, for FluxEdit, both $\bar{A}_{ZP}$ and $\bar{A}_{ZZ}$ can be used to generate masks, with $\bar{A}_{ZP}$ producing clearer masks at earlier steps. For Omnigen, however, only $\bar{A}_{ZZ}$ is suitable for mask generation.

\begin{figure*}[h]
\centering
\includegraphics[width=0.9\linewidth]{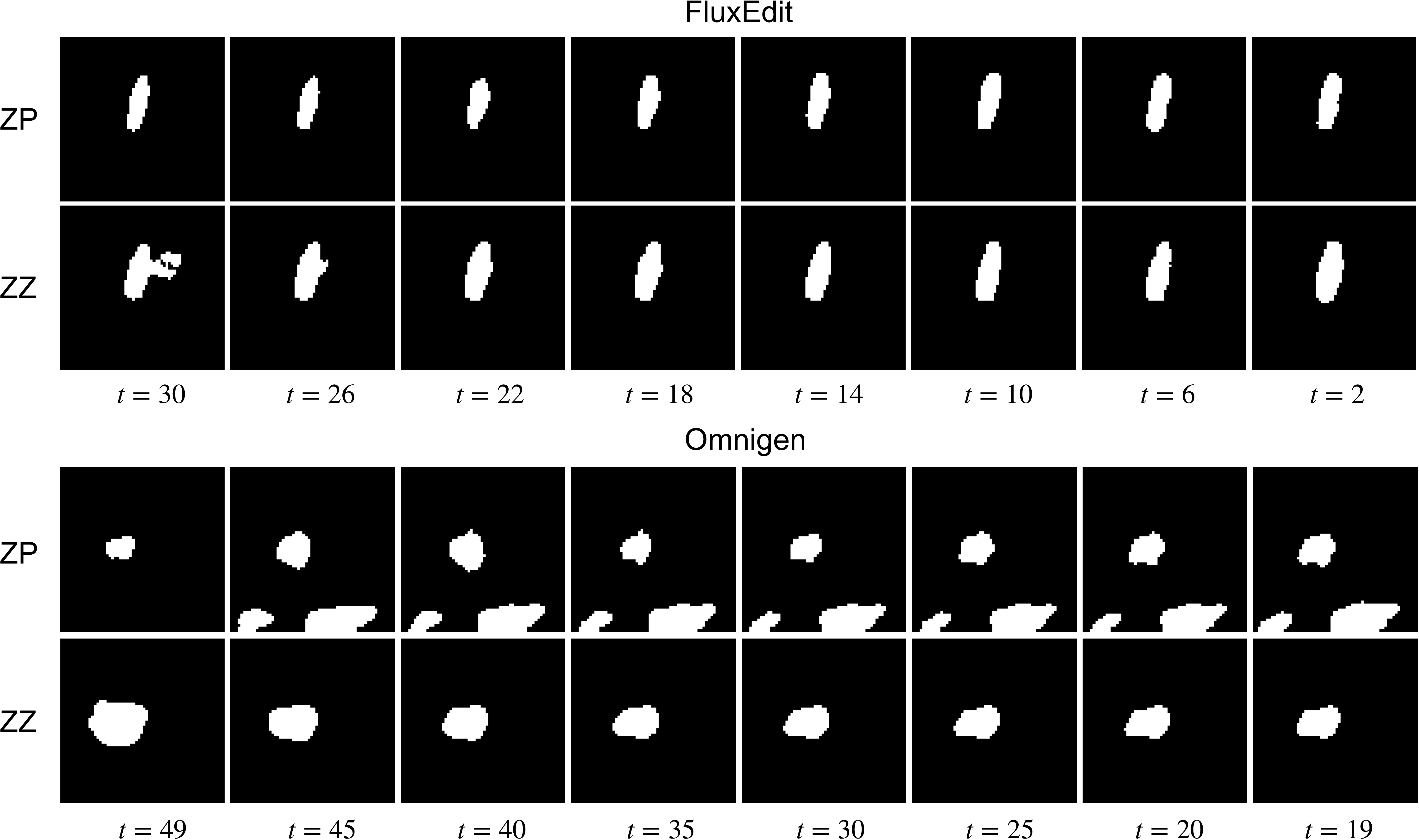}
\caption{Ablation study on the influence of the step used for mask extraction on the quality of the generated masks.}
\label{fig6}
\end{figure*}

\begin{figure*}[h]
\centering
\includegraphics[width=0.9\linewidth]{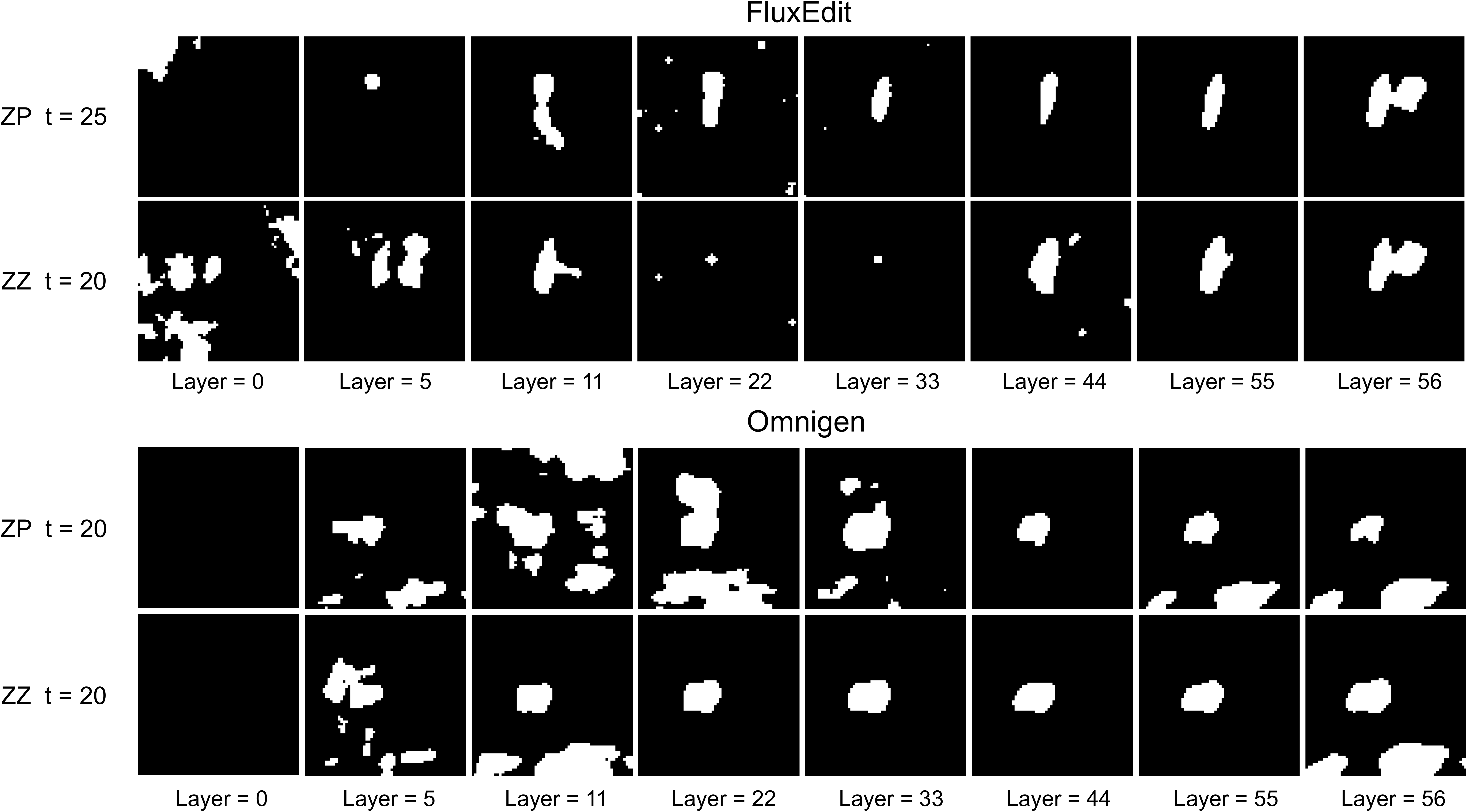}
\caption{Ablation study on the influence of the layer used for mask extraction on the quality of the generated masks.}
\label{fig7}
\end{figure*}

\begin{figure*}[h]
\centering
\includegraphics[width=0.9\linewidth]{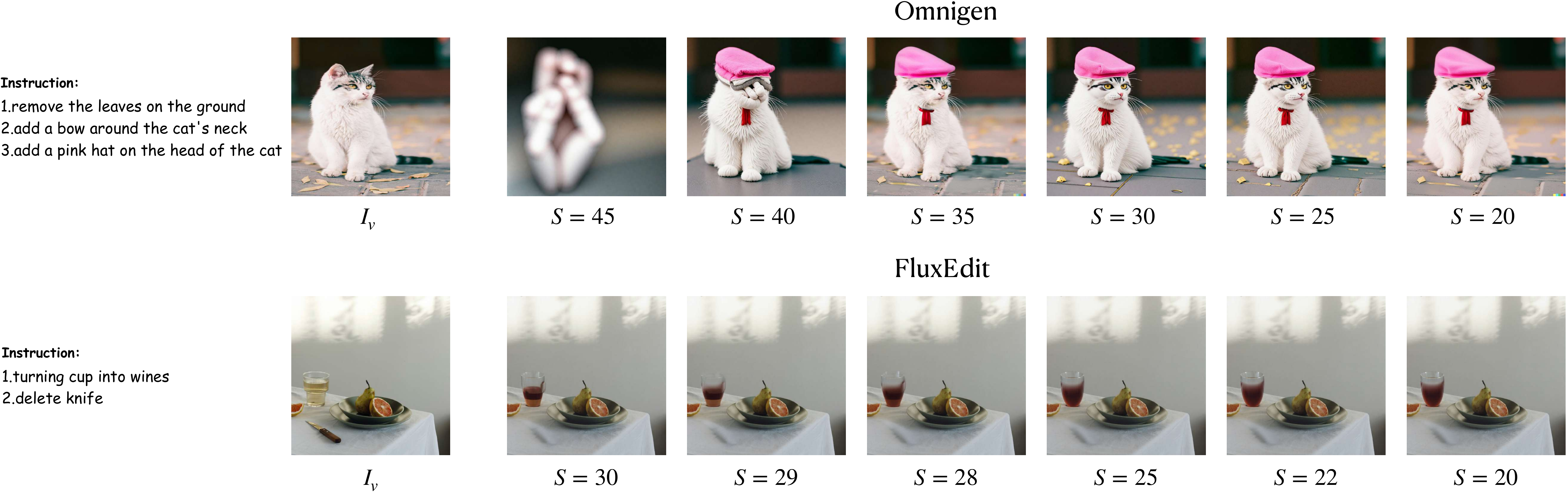}
\caption{Ablation study on the effect of the pre-defined step $S$ used for multi-instruction disentanglement on the quality of editing. The top row of images corresponds to OmniGen, while the bottom row corresponds to FluxEdit.}
\label{fig8}
\end{figure*}

\section{Ablation study on Mask Generation}
\label{sec:app2}
\noindent\textbf{Ablation of Timesteps.} As demonstrated in Fig. \ref{fig6}, we extract masks from the penultimate layer of both editing models at different timesteps. Generally, as the diffusion steps progress, the quality of the masks improves. After a certain number of steps, the changes in the masks become minimal, and the results stabilize.

\noindent\textbf{Ablation of Extracted Layers.} As illustrated in Fig. \ref{fig7}, we extract masks for each instruction from different layers of the editing models. The results indicate that for both editing models, lower layers are not suitable for mask extraction, as information tends to interact across the image, resulting in dispersed attention weights. As the layer depth increases, the semantic information of different tokens gradually converges to form higher-level semantic features, enabling a more focused attention on the edited regions. Notably, the masks extracted from the final layer exhibit lower quality. We suggest that this may be because the feature of the final layer is utilized for image reconstruction through the VAE, making it less effective for mask generation. A similar phenomenon has been observed in transformer-based text classification models, where the final layer features are not always ideal for semantic representation \cite{DBLP:journals/corr/abs-2406-11890}, and the textual features also demonstrate a hierarchical pattern \cite{DBLP:conf/acl/JawaharSS19}. Therefore, we use the penultimate layer to extract masks in our main experiments.
\section{Ablation study of Pre-defined Step}
\label{sec:app3}
To investigate the influence of the pre-defined step $S$ on the quality of edited images, we test FluxEdit and OmniGen using different timesteps $S$. As shown in Fig. \ref{fig8}, for OmniGen, when $S \geq 45$, the generated images deviate significantly from both the original image and the editing instructions. When $S = 40$, the generated images fulfill the instructions but compromise the consistency of the unedited regions with the original image. When $S \leq 20$, the generated images perform well in both fulfilling the instructions and maintaining consistency with the original image.
We suggest that this behavior is due to OmniGen's unique attention mechanism, as defined in Eq. \ref{eq:5} of the main text. Specifically, when $t < T$, where $T$ is the total number of timesteps, there is continuous interaction between the text tokens and the noise image tokens, but no further interaction involving the text tokens. We argue that this causes the noise image tokens to contain significant semantic information from the instructions. When these latent image tokens are blended using masks, the semantic information of the instructions is disrupted, leading to the poor results seen in the top row of images in Fig. \ref{fig8} when $S = 45$. Therefore, for OmniGen, we set $S = 15$ to prevent the disruption of instruction-related information.
Moreover, for FluxEdit, it is feasible to perform multi-instruction disentanglement in the early steps of reverse diffusion possibly because mask-based latent blending will not destroy the original feature space in these steps. As a result, we set $S = 27$ for FluxEdit.

\end{document}